\documentclass[10pt, two column, twoside]{IEEEtran}
\usepackage{color}
\usepackage[colorlinks, linkcolor=color1, anchorcolor=blue, citecolor=color1]{hyperref}
\hypersetup{
    urlcolor=blue,
}

\usepackage{amssymb}
\usepackage{amsmath}
\usepackage[lined,boxed,commentsnumbered, ruled]{algorithm2e}
\usepackage{mathrsfs}
\usepackage{algorithmic}
\usepackage{bm}
\usepackage{cite}

\usepackage{pgfplots}
\usepackage{tikz}
\usetikzlibrary{arrows}
\usepackage{subfigure}
\usepackage{graphicx,booktabs,multirow}

\definecolor{colorhkust}{RGB}{20,43,140}
\definecolor{colortsinghua}{RGB}{116,52,129}
\definecolor{color1}{RGB}{128,0,0}

\date{}

\begin{document}

        \title{Communication-Computation Trade-Off in Resource-Constrained Edge Inference}
\author{Jiawei Shao, Jun Zhang
        
        \thanks{Jiawei Shao and Jun Zhang are with the Department of Electronic and Information Engineering, The Hong Kong Polytechnic University, Hong Kong (e-mail: jiawei.shao@connect.polyu.hk, jun-eie.zhang@polyu.edu.hk).}
                }

\maketitle

\maketitle

\begin{abstract}
The recent breakthrough in artificial intelligence (AI), especially deep neural networks (DNNs), has affected every branch of science and technology.
Particularly, edge AI has been envisioned as a major application scenario to provide DNN-based services at edge devices.
This article presents effective methods for edge inference at resource-constrained devices. It focuses on device-edge co-inference, assisted by an edge computing server, and investigates a critical \emph{trade-off} among the computational cost of the on-device model and the communication overhead of forwarding the intermediate feature to the edge server.
A general three-step framework is proposed for the effective inference: (1) model split point selection to determine the on-device model, (2) \textit{communication-aware} model compression to reduce the on-device computation and the resulting communication overhead simultaneously, and (3) task-oriented encoding of the intermediate feature to further reduce the communication overhead.
Experiments demonstrate that our proposed framework achieves a better trade-off and significantly reduces the inference latency than baseline methods.

\end{abstract}

\section{Introduction}

Recent advancements in Deep Neural Networks (DNNs) have led to successful applications in a broad spectrum of domains, including computer vision, speech recognition, and natural language processing \cite{krizhevsky2012imagenet}.
Driven by the demand for deploying DNN-based services at various edge devices (e.g., smartphones, wearables, IoTs), a new research area called \emph{edge AI} emerges \cite{zhu2020toward}.
Edge AI consists of edge training, i.e., to train DNN models based on data distributed at different devices, and edge inference, i.e., to provide DNN-based execution at resource-constrained devices.
While communication-efficient methods for edge training have received significant attention \cite{zhou2019edge}, the counterpart on edge inference is less well investigated. This article aims to fill this gap and introduces new design problems and methodologies for edge inference by presenting a delicate trade-off between communication overhead and on-device computational cost.

Currently, the status quo of edge inference is either execution on the mobile devices (\textit{on-device inference}) or offloading to the edge server for execution (\textit{server-based inference}). Unfortunately, on-device inference provides limited accuracy due to the constrained resources.
Some works applied model compression \cite{cheng2017model_compression_survey} to alleviate the huge computational complexity, relying on the fact that DNNs are typically over-parametrized.
Nevertheless, it is challenging to achieve high accuracy with compact models, e.g., there is a roughly 10\% gap between the state-of-the-art mobile model and the best model \cite{cai2019once}.
On the other hand, server-based inference induces excessive communication overhead, making it challenging to support latency-sensitive applications like self-driving cars. Moreover, it also suffers from the data privacy issue. Thus, the limited computation resource and the excessive communication overhead form bottlenecks for on-device inference and server-based inference.

\begin{figure}[t]
\centerline{\includegraphics[width=0.45\textwidth]{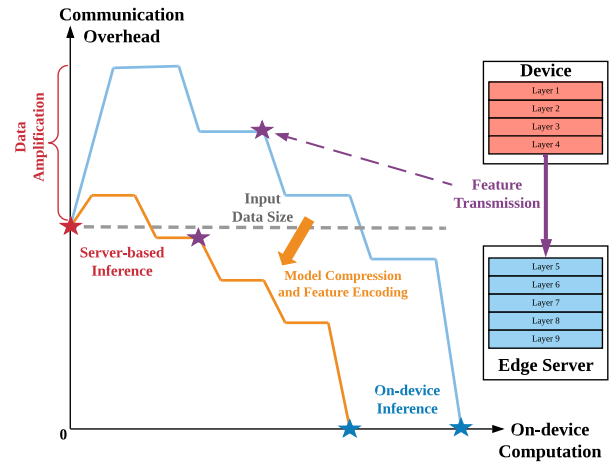}}
\caption{The communication-computation plane of edge inference in ResNet \cite{ResNet} for classification tasks. \textbf{(1) Trade-off:} the blue and orange curves correspond to the on-device computation and communication overhead at different split points, where the blue curve corresponds to the original network, and the orange curve corresponds to the network after model compressed and feature encoding.
\textbf{(2)  Data amplification:} As suggested in \cite{JALAD}, the data amplification means the communication overhead of the intermediate feature is larger than that of input data. The grey dashed line is the communication overhead of input data.
\textbf{(3) Special points:} the red and blue stars correspond to on-device inference and server-based inference. The purple star corresponds to one case of device-edge co-inference. 
With model compression and feature encoding, the on-device computation and communication overhead is reduced, and the data amplification effect is alleviated.}
\label{trade-off1}
\end{figure}

\begin{figure*}[t]
\centerline{\includegraphics[width=0.891\textwidth]{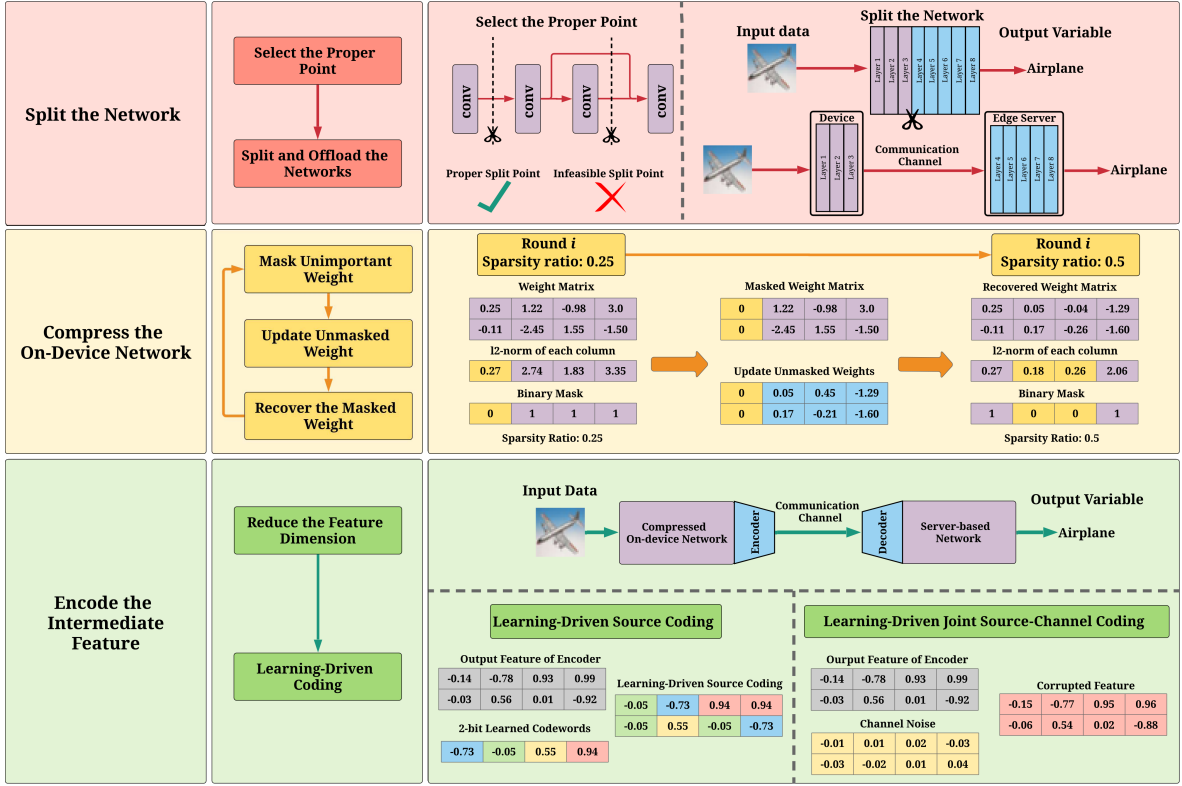}}
\caption{The proposed framework of device-edge co-inference. 
\textbf{(1) Split the Network:} The input of the framework is the pre-trained DNN. The first step is to select the split point to divide the DNN into two parts. The front part of the neural network is deployed on the edge device, and the other part is offloaded on the edge server. 
\textbf{(2) Compress the on-device model:} The on-device model is compressed by incremental network pruning. In each iteration, the mask would remove the unimportant weights (set their value to 0) based on their $l_{2}$-norm. Then the unmasked weights are updated in the back-propagation. After that, the masked weight would be recovered, and the next iteration starts.
In the training process, the sparsity ratio will continuously increase until it reaches the desired ratio.
\textbf{(3) Encode the intermediate feature:} With the compressed on-device model, we use a pair of lightweight encoder-decoder structure to shrink the volume of the intermediate feature. Besides, using learning-driven source coding or joint source-channel coding, we further reduce the communication overhead by learning the mapping from each symbol to codeword.}
\label{edge-inferece-arch}
\end{figure*}

The device-edge co-inference method effectively removes the computation and communication bottlenecks mentioned above. It splits a large DNN into two parts: a computation-friendly model is deployed on the edge device, while the other part, usually with a larger size and heavy computation, is deployed on the edge server.
The split point selection is critical, which affects the resulting on-device computational cost and communication overhead. 
Although the entropy of the intermediate feature decreases layer by layer in the supervised learning \cite{tishby_ib_2015}, simply selecting the split point can hardly reduce the inference latency due to the \emph{data amplification} effect \cite{JALAD}, i.e., the intermediate output data is larger than the input data. Fig. \ref{trade-off1} show such an effect based on ResNet \cite{ResNet}.

This article investigates the communication-computation trade-off in device-edge co-inference and proposes a general three-step framework shown in Fig. \ref{edge-inferece-arch} to reduce the co-inference latency in supervised tasks.
Built upon a rudimentary trade-off via split point selection, we introduce effective techniques to reduce the on-device computation and communication costs.
The first technique, named \emph{communication-aware model compression}, is motivated by the recent development in DNN model compression \cite{DBLP:conf/iclr/ZhuG18}. It compresses the on-device model to reduce the computational cost while accounting for the induced communication overhead by controlling activated neurons.
The second technique, named \emph{task-oriented feature encoding}, leverages the powerful DNN as a learning-driven encoder \cite{NJSCC} to compress the output feature of the on-device model.
By reducing the communication overhead and alleviating the data amplification effect, feature coding also makes it possible to split at earlier layers, leading to a reduction of on-device computation, as illustrated in Fig. \ref{trade-off1}.

This framework can be applied to general supervised tasks. It exploits the fact that the intermediate feature has a lower entropy compared with the input data \cite{tishby_ib_2015}, which makes it possible to reduce the transmission latency. 
Besides, the adopted model compression and feature encoding methods are not limited to a particular case, but can be applied to more general tasks \cite{NJSCC,deniz_jssc,DBLP:conf/iclr/ZhuG18,2-step}.

Our main contributions are summarized as:
\begin{itemize}
\item We propose a general three-step framework to strike a communication-computation trade-off caused by the local computation workload and communication overhead.
\item We adopt an incremental network pruning method to reduce the redundant weights and computation latency while alleviating the communication overhead.
\item We use a two-step feature encoding to reduce the communication latency by learning a compact representation.
\item We conduct experiments based on the classification task that evidence the proposed framework's advantages in striking a better communication-computation trade-off and reducing the edge inference latency.
\end{itemize}

The following article is organized as follows. In Section II, we illustrate the communication-computation trade-off in co-inference and the general three-step framework. Then details of communication-aware model compression and task-oriented feature encoding are presented in Section III and Section IV. The experiments are conducted in Section V based on the classification task. 
Finally, Section VI concludes this article and points out future directions.

\section{Communication-Computation trade-off in Edge Inference}
\label{section2}

This section first presents communication-computation trade-off as a critical design perspective in device-edge co-inference and introduces split point selection as a rudimentary approach to achieve a baseline trade-off. 
Then we present two perspectives to improve the trade-off for more effective edge inference. 
Finally, we propose a general framework as shown in Fig. \ref{edge-inferece-arch} that incorporates the three techniques.

\subsection{Communication-Computation Trade-Off and Model Splitting}
\label{split point}

Limited on-device resources fundamentally constrain the performance of edge inference. Many devices are with limited computing capability due to low-end processing units and limited communication capability caused by the limited bandwidth.
Device-edge co-inference is a promising method to overcome these limitations for efficient inference, which splits a DNN into two parts, to be deployed at a device and an edge server.
As the DNN gradually abstract the intermediate feature \cite{tishby_ib_2015} in supervised tasks, the low-latency edge inference can be achieved by carefully selecting the split point to control the on-device model size and its output dimension.

Split point selection provides a rudimentary way for communication-computation trade-off.
For example, Neurosurgeon \cite{kang2017neurosurgeon} was proposed as a co-inference system to reduce the end-to-end latency while satisfying energy and memory constraints via split point selection. 
However, the split-only method achieved limited latency gains due to the data amplification effect.
Considering the redundancy in model weights \cite{DBLP:conf/iclr/ZhuG18} and intermediate features, we introduce model compression and feature encoding in the following two subsections for further latency reduction, with more details provided in Sections \ref{section3} and Section \ref{section4}.

\subsection{Communication-Aware Model Compression for Better Trade-Off}

The over-parameterized DNN hinders its deployment on resource-constrained devices.
Model compression has been a widely applied technique to reduce the model size and accelerate the execution with little performance loss thanks to the parameter redundancy. 
In device-edge co-inference, we prune the unnecessary parameters in the on-device network.
Meanwhile, considering that the on-device network output affects the communication latency, the compression method should be \emph{communication-aware}.
Structured pruning is a suitable method, which removes a regular group of weights and realizes hardware acceleration (e.g., channel-wise pruning) and output data size reduction (e.g., neuron-wise pruning).

There have been edge-based methods that deployed large models on resource-constrained devices by aggressive compression or knowledge distillation. However, the over-compressed model may lead to intolerant performance loss.
For edge-device co-inference, the 2-step pruning method \cite{2-step} used the iterative pruning workflow to compress the whole neural network. However, compressing the server-based network provides limited latency gains given the abundant computation resource at the server, and it may reduce the compression capability of the on-device network.
In Section \ref{section3}, we introduce the incremental network pruning method that is a kind of magnitude-based pruning method \cite{DBLP:conf/iclr/ZhuG18}, serving as a communication-aware model compression method.

\subsection{Task-Oriented Feature Encoding for Better Trade-Off}
As shown in Fig. \ref{trade-off1}, there is a data amplification effect \cite{JALAD} in DNN.
Although communication-aware model compression can alleviate this problem to some extent, it does not take the sparsity in the feature maps into account, i.e., the data size is far away from its entropy.
Besides, the feature transmitted to the edge server should only contain a \emph{minimal} amount of information that is \emph{sufficient} for the inference task.
These two factors motivate the emergence of task-oriented feature encoding to get a compressed representation. Particularly, the feature encoder is implemented by a lightweight DNN that is trained together with other layers.

Many related works, e.g., BottleNet++ \cite{iccshao}, JALAD \cite{JALAD}, utilized traditional coding methods like Huffman coding and JPEG for edge-device co-inference.
The main difference between the task-oriented encoding and traditional coding techniques is that the traditional methods aim to reconstruct the data in the receiver. However, task-oriented encoding allows discarding plenty of information unrelated to the task.
In Section \ref{section4}, we introduce more details about the proposed two-step feature encoding.

\subsection{Framework Overview}
\label{framework}
Based on the above discussion, we propose a general framework shown in Fig. \ref{edge-inferece-arch}, which combines three techniques, namely model splitting, model compression, and feature encoding.
The workflow consists of three steps.
First, we choose a split point that does not (highly) suffer from the data amplification effect.
Second, we use the model compression method to prune the redundant weights incrementally. In this process, we continuously increase the sparsity ratio until it reaches the performance (e.g., accuracy) threshold.
Third, we use the feature encoding to compress the intermediate data with acceptable performance degradation.
Note that there is a tight coupling among split position, model sparsity ratio, and data compression ratio, and it is hard to choose these three parameters automatically 
under the particular edge environment.
In our investigation, we manually tune the parameters through a brute force search. All the output models are profiled and stored in a lookup table.
When we need to deploy a DNN to the network edge, we can choose the best model from the table for the particular edge environment.
More details about the model compression and feature encoding are introduced in the next two sections.

\section{Communication-aware Model Compression}
\label{section3}

This section first introduces conventional methods for DNN
model compression. Then, methods that can be applied for
communication-aware model compression are identified. Implementation details about a particular method, incremental network pruning, are presented next.

\subsection{Model Compression for DNNs}

\label{compression related work}
There have been lots of interests in DNN model compression \cite{cheng2017model_compression_survey}, which effectively reduces the memory footprint and computational cost of DNNs.
These methods can be categorized into four schemes: \emph{parameter pruning},  \emph{low-rank factorization}, \emph{compact convolutional filters}, and \emph{knowledge distillation}.
Generally, many works on edge inference \cite{2-step} applied parameter pruning as it is robust and flexible. These parameter pruning methods have less constraints on the network structure and limited extra computational cost.
In contrast, low-rank factorization is hard to implement because the decomposition operation is computationally expensive. 
Compact convolutional filters apply strong transform constraints to convolutional filters, which achieve on-pair performance in wide DNNs but not in deep networks \cite{ResNet}.
For knowledge distillation based methods, although they straightly reduce the model size by re-training a shallower model, they are only applicable to tasks with the Softmax loss function \cite{cheng2017model_compression_survey}, and the re-training process is time-consuming.
In the next subsection, we identify a suitable pruning method that can be applied for communication-aware model compression.

\subsection{Communication-Aware Model Compression}

Model compression for device-edge co-inference should account for both the on-device computational cost and the resulting communication overhead.
Thus, it should be designed in a \textit{communication-aware} manner, and directly applying existing compression methods may not be effective.
Parameter pruning is a good candidate. 
There are two categories of pruning: \emph{unstructured} pruning and \emph{structured} pruning \cite{DBLP:conf/iclr/ZhuG18}. 
The unstructured pruning aims to prune individual unimportant weight and can achieve a high sparsity ratio. However, the pruned weight matrices with irregular sparsity may not be helpful for matrix multiplication acceleration. So, most previous works used structured pruning to accelerate the DNN by obtaining regular sparsity pattern (e.g., channel-wise pruning, kernel-wise pruning).
Our framework follows previous works such as AutoML, and uses the structured pruning method to compress the DNN.
Besides, we use the activation pruning to prune the individual neuron in the last layer of the on-device network to reduce the communication overhead.

\subsection{Incremental Network Pruning}%network pruning

In our framework, we use incremental network pruning, as shown in the yellow region of Fig \ref{edge-inferece-arch}.
It is a magnitude-based \emph{pruning during training} method similar to \cite{DBLP:conf/iclr/ZhuG18}.
Its advantage compared with \emph{one-shot pruning} and \emph{pruning before training} methods is that it can reach a high compression ratio with limited performance loss.
Its iterative pruning process in training is explained below.

We first set targeted channel-wise sparsity ratios at the beginning of each pruning iteration.
Particularly, in the output layer, a high sparsity ratio means high communication overhead reduction.
There is a binary mask for every layer of the on-device model.
The mask with value 0 means that the corresponding output channel will be masked (pruned) in this iteration.
The pruning process consists of three steps in each iteration.
(1) We sort the weights corresponding to each output channel based on their $l_{2}$-norm.
The smallest $S_{i}\%$ of weights will be masked in this iteration, where $S_{i}\%$ is the corresponding sparsity ratio.
(2) In the forward-propagation, the masked weights' values are set to 0, and only the unmasked weights will be updated in the back-propagation.
(3) The mask is deleted, and masked weights are recovered to their original values at the beginning of this iteration. Then our method starts the next iteration.
The sparsity ratio in each iteration will continuously increase until it reaches the desired ratio, and this mechanism makes the pruning process stable.

\section{Task-oriented Feature Encoding}
\label{section4}

While communication-aware model compression can partly reduce the communication overhead, its capability is limited due to the high dimension-wise correlation.
In this section, we propose task-oriented feature encoding to reduce the communication overhead further.
In the following, we first elaborate the critical difference between coding in traditional communication system and that for tasked-oriented compression.
Next, we introduce the implementation details of the proposed two-step feature encoding.

\subsection{Communication-Oriented vs. Task-Oriented Encoding}
%Communication-oriented coding
There are many hand-crafted coding methods to compress the data in traditional communication, including source coding (e.g., Huffman coding, PNG, JPEG) and joint source-channel coding.
Besides, recent works applied DNNs in feature encoding for data transmission, e.g., NECST \cite{NJSCC} and DeepJSCC \cite{deniz_jssc}.
These communication-oriented methods aim to recover the data, either perfectly (lossless coding) or with tolerable distortion (lossy coding). 
Compared with these methods, task-oriented encoding can achieve latency reduction for two reasons.
First, data recovery is unnecessary in task-oriented encoding, and unnecessary information can be discarded, which adapts to the feature distribution and is trained based on a target objective.
Besides, due to DNNs' fault-tolerant property, although the received data is corrupted by over-compression or channel noise, the performance would not severely degrade \cite{iccshao}.
Note that as the task-oriented feature encoding is highly related to the particular task, it cannot be used as a general encoding method, such as Huffman coding, in the communication.
The next subsection introduces the proposed \emph{two-step pruning} method adopted in our framework.

\subsection{Two-Step Feature Encoding}
We propose a two-step encoding method, consisting of (1) data dimension reduction to shrink the volume of the intermediate feature and (2) learning-driven coding to further compress the feature by mapping each symbol to the codeword.
Both steps are implemented via (extra) neural networks, which are trained in an end-to-end manner for maximizing the inference accuracy of the task.
In Fig. \ref{edge-inferece-arch}, the intermediate feature encoding is shown in the green region of the framework.

\subsubsection{Dimension Reduction}

Inspired by the recent work using the auto-encoder structure for communication \cite{iccshao}, we use a pair of lightweight complementary encoder and decoder (with less than 1\% additional computational cost compared with the original network) to reduce the data dimension.
Previous work \cite{DBLP:conf/iclr/ZhuG18} indicated the existence of redundant channels in the intermediate feature.
In our experiment, the first layer in the encoder is a convolutional layer to reduce the unnecessary channels.
Next, the reshaped feature passes a fully connected layer to further reduce its dimension.
At the edge server, the decoder reconstructs the data by a complementary structure.

\subsubsection{Learning-Driven Coding}

In the second step, the DNN learns the mapping from source symbols to codewords.
We provide two alternative coding methods: learning-driven source coding and learning-driven joint source-channel coding (JSCC).
The former method should be implemented with a traditional channel encoder to combat the channel noise, while the latter method models the noisy channel in the training process.

\textbf{Learning-Driven Source Coding:}
As the original float-point numbers in the intermediate features induce a significant communication overhead, using (lossy) source coding via quantizing the float-point numbers to fewer bits can considerably reduce the overhead.
For the learning-based method, the DNN learns a set of optimal (discrete) codewords for the lossy compression, rather than using traditional methods like rounding or truncation with pre-deﬁned codewords. These learning-driven codewords are adaptive to the input data distribution and can minimize the encoded information loss.

\textbf{Learning-Driven Joint Source-Channel Coding:}
\label{jssc}
These coding schemes model a noisy channel as a non-trainable layer in the DNN, which can reduce the redundancy compared with the separate channel coding for two reasons. First, the DNN-based coding scheme can find better codeword mapping than the traditional channel coding, e.g., LDPC and Turbo code. Besides, due to DNNs' fault-tolerant property, although the channel noise corrupts the transmitted data, it has relatively less effect on edge inference performance.

\section{Experiment and Evaluation}
\label{section6}

%\label{AIapp}
%With 

\subsection{Experimental Setup}

\label{experiment}

\subsubsection{Dataset and Neural Network}
We evaluate our proposed framework on an image classification task with the CIFAR-10 dataset \cite{cifar100}, which consists of 60,000 32$\times$32 color images in 10 classes.
We use the classical ResNet18 \cite{ResNet} as a classifier, which is trained on the 50,000 training images and evaluated on the 10,000 testing images.

\subsubsection{Metric}
For this DNN-based application, one important metric is classification accuracy. Our result shows that the original ResNet18 can reach around 95\% accuracy, and we set the accuracy threshold to 93\% in edge inference.
Another important metric is the end-to-end latency, which is mainly influenced by the on-device computation and communication overhead.
The on-device computation workload is approximated by the float-point operations (FLOPs), and the communication overhead is evaluated by the transmitted data size (bits) or transmitted latency (seconds).

\subsubsection{Baselines} To the best of our knowledge, there has been no systematic study of the communication-computation trade-off in device-edge co-inference. To fully illustrate this trade-off and verify the effectiveness of our framework, we consider three baselines: 
(1) the original network with split point selection, but without model/feature compression; (2) the 2-Step Pruning method \cite{2-step}, where the first step prunes the entire DNN to shrink the on-device model and the second step only prunes the layer right before the split point to reduce the communication overhead; (3) BottleNet++ \cite{iccshao}, which uses an auto-encoder structure to compress the intermediate data. 
As BottleNet++ and our method using DNNs to encode the feature, Huffman coding is used in the original network and the 2-Step Pruning method for a fair comparison. Besides, we use the PNG method for input image compression, corresponding to the server-based method's communication overhead.
The code is available at \url{https://github.com/shaojiawei07/Edge_Inference_three-step_framework}.

\subsection{Communication-Computation Trade-Off}

\label{exp1}
\begin{figure}[t]

\centerline{\includegraphics[width=0.47\textwidth]{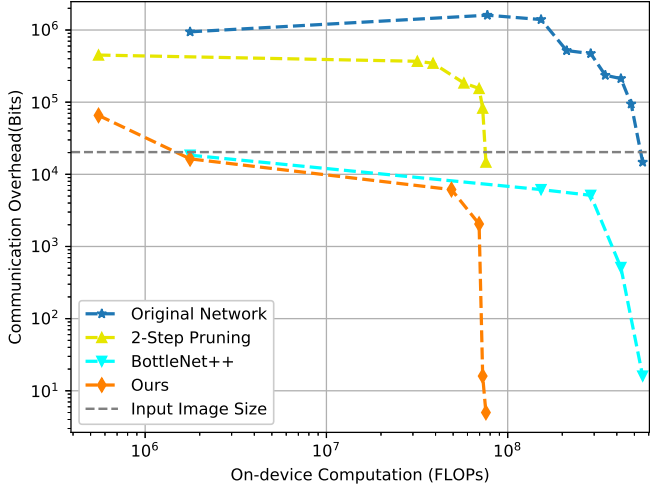}}

%\caption{}

\caption{The communication-computation trade-off curves in device-edge co-inference.}
\label{trade-off}
\end{figure}

Fig. \ref{trade-off} plots the communication-computation trade-off curves of our method and three baselines.
First, for each point on our framework's curve, there is no point on other curves with smaller computational cost and communication overhead simultaneously.
Our framework achieves a significant saving in both the on-device computational cost and communication overhead compared with the original network.
So, our method can enable effective inference on resource-constrained devices and is applicable in varying combinations of device computation capability and available bandwidth.

Next, we analyze the four curves from the data amplification perspective \cite{JALAD}.
The grey dashed line is the communication overhead for transmitting the input data.
Nearly all the original network's intermediate features and that of the 2-Step Pruning method are larger than the input data.
Although 2-Step Pruning considers reducing the communication overhead in the second pruning step, simply removing the redundant channel is not sufficient to alleviate the data amplification effect.
So they have few available split points for device-edge co-inference, and these two settings can hardly reduce the end-to-end latency.
For BottleNet++, it has a great potential to compress the intermediate feature size, and all the communication overheads at the split points are smaller than the input data.
However, BottleNet++ does not compress the on-device model, so most of the split points suffer excessive on-device computation, which will increase the inference latency and energy consumption.
Our method makes a better communication-computation trade-off, which remedies the influence of data amplification and requires less on-device computation.

Note that, in the low on-device computation region, Fig. \ref{trade-off} shows a marginal performance gain of the proposed method because the limited computation resources (corresponding to few deployed layers) cannot abstract the input data to a low-entropy representation.
Besides, a lightweight auto-encoder structure cannot produce a compact representation given the large volume of the intermediate feature caused by the data amplification.
This indicates the importance of neural network architecture design for edge AI, which is a promising future research direction.

\subsection{Edge Inference Speedup}
\label{exp2}
This part provides a real-world case study to compare the end-to-end latency of different methods.
We select the Raspberry Pi 3 as the edge device (1GB RAM and 24GFLOP/s), and the edge server is with RTX 2080 TI (11GB RAM and 13.45 TFLOP/s).
Note that the Raspberry is constrained by its memory resource because its CPU and GPU share the 1GB RAM, and other modules also occupy the memory.
We use Pytorch to build the ResNet18, and the original network needs more than 0.9 GB memory.
We can roughly deploy 40\% of the uncompressed model and 90\% of the compressed model on the Raspberry due to the memory constraint.

\begin{figure}[t]

\centerline{\includegraphics[width=0.47\textwidth]{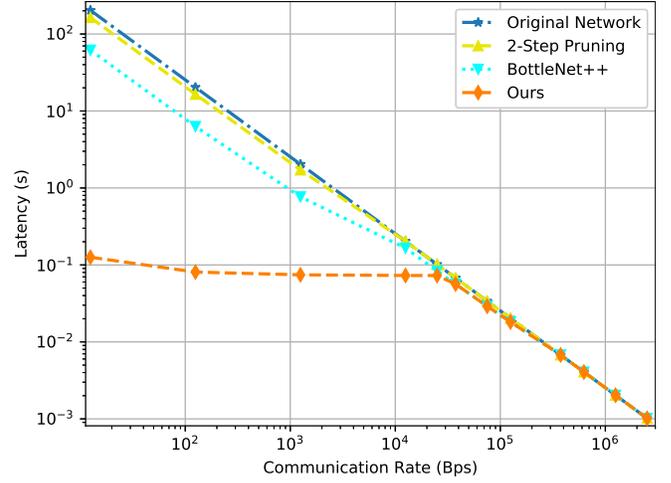}}

\caption{Latency as a function of communication rate. Our method can maintain the latency around 0.1s when the communication becomes poor (less than 40KB)}
\label{latency_communication}
\end{figure}

With the fixed computation capability and memory constraint, we test the inference latency under different communication rates.
In Fig. \ref{latency_communication}, we observe that when the edge device under a poor communication environment (smaller than 40KBps), our method can maintain the inference latency around 0.1s.
However, for the three baselines, the end-to-end latency increases dramatically.
The reason is that our method, with more available split points, can flexibly adjust the number of on-device offloading layers and easily strike a balance between the on-device computation and communication overhead.
Thus, our method can work effectively in resource-limited scenarios.

\begin{figure*}[h]
\subfigure[]{
\begin{minipage}[]{0.47\linewidth}
\centering
\includegraphics[width=1\textwidth]{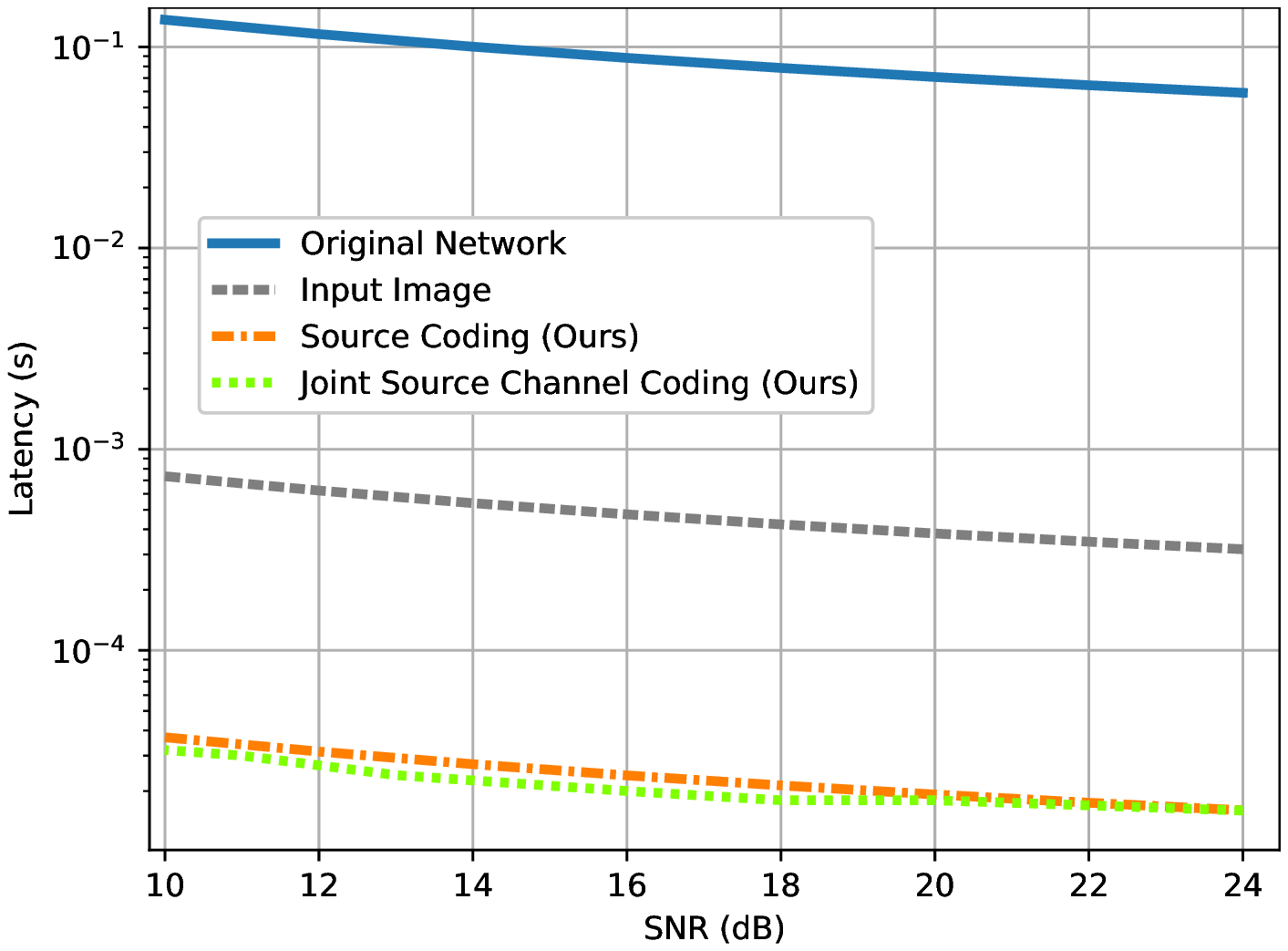}
\label{AWGN}
%\caption{Latency as a function of SNR.}
\end{minipage}%
}%
\subfigure[]{
\begin{minipage}[]{0.47\linewidth}
\centering
\includegraphics[width=1\textwidth]{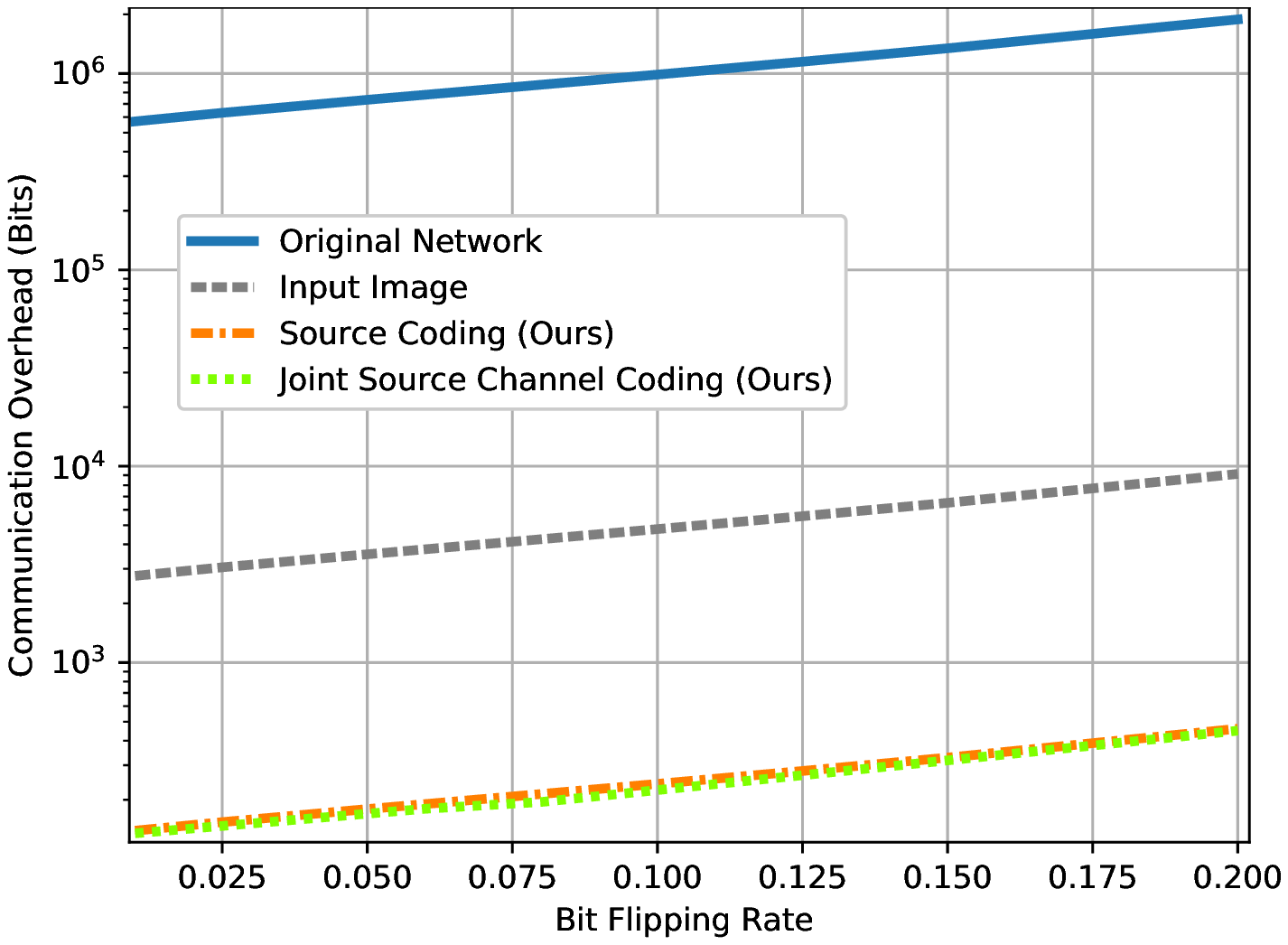}
\label{BSC}
%\caption{Communication Overhead as a function of bit flipping rate.}
\end{minipage}%
}%
\centering
\caption{(a) The communication latency under different SNR's in the AWGN channel with channel bandwidth $W=1\textup{MHz}$ and (b) the communication overhead under different bit flipping rates in the Binary Symmetric Channel.}
\label{jscc}
\end{figure*}

\subsection{Performance of Learning-Driven JSCC over Noisy Channels} % coding method
%set accuracy threshold
\label{experiment_jssc}
\label{exp3}
Previous simulations assumed error-free transmission of the intermediate feature. In this part, we consider noisy channels and verify the effectiveness of the learning-driven JSCC scheme for feature encoding.

In this experiment, we split the neural network behind the \emph{Conv\_4x} \cite{ResNet}.
Then we compare the communication overhead of learning-driven coding (assuming perfect channel coding) and learning-driven JSCC.
Besides, to show the performance gain, we also compare with the communication overhead of (1) the input data and (2) the intermediate feature of the original network.
We consider two kinds of noisy channels, namely, an AWGN channel and a Binary Symmetric Channel (BSC), where the noise is characterized by the signal-to-noise ratio (SNR), and bit flipping rate $p$, respectively.
Note that JSCC does not need extra channel coding to protect the signals/bitstream against channel noise, and we assume perfect channel coding for other methods.
Specifically, we adopt the Shannon capacity bound as the communication rate for baseline methods, so we are comparing with the performance upper bounds of these methods.

Fig. \ref{jscc} shows that the communication overhead of JSSC is slightly lower than or comparable with learning-driven source coding with optimal channel coding and outperforms the other two baselines by a large margin.
Besides, JSSC is with much lower encoding/decoding complexity, as it is constructed by the DNN. 
This comparison illustrates the great potential of JSCC in edge inference, as it is able to reduce the communication overhead, processing time and energy.

\section{Conclusions and Future Works}
\label{section7}
This article investigate the communication-computation trade-off in edge inference and propose a general three-step framework for supervised tasks, which incorporates three techniques, i.e., model splitting, model compression, and feature encoding to reduce the end-to-end latency in co-inference.
Simulations based on the classification task verify that this framework achieved a better communication-computation trade-off and much lower latency than other baselines.

While promising results have been demonstrated, the proposed framework also has some limitations.
First, the incremental network pruning method requires manual setup for the sparsity ratio in each iteration, and the \emph{pruning during training} approach requires many iterations to coverage.
Second, the feature encoding scheme uses extra DNNs that introduce an additional computational cost. Designing a lightweight DNN that maintains a trade-off between extra computation and the data compression ratio becomes essential. Besides, feature encoding should also settle the potential performance loss in discretization.
Third, the hyperparameters in the framework (split position, model sparsity ratio, and data compression ratio) are manually tuned, and a brute force search is used to find the candidate model in different edge environments. This off-line search process is time-consuming due to the large search space.

The above limitations bring abundant research opportunities.
First, the neural architecture search (NAS) technique \cite{cai2019once} can be applied for searching the hardware-accelerated and communication-aware DNNs for edge deployment, rather than the hand-crafted models.
Second, to further reduce the communication overhead, information theory can be leveraged to design the feature encoder. 
For example, we can add an entropy constraint (for source coding) and a mutual information constraint (for joint source-channel coding) during the network training.
Third, following AutoML, searching the optimal hyperparameters by reinforcement learning (RL) is a promising direction to explore. The search result could be adaptive to different edge environments.

{\bf{Jiawei Shao}} [S'20] (jiawei.shao@connect.polyu.hk) received his B.S. degree from Beijing University of Posts and Telecommunications. He is currently pursuing a Ph.D. degree at The Hong Kong Polytechnic University. \\[-3mm]

{\bf{Jun Zhang}} [S'06-M'10-SM'15] (jun-eie.zhang @polyu.edu.hk) received his Ph.D. degree from the University of Texas at Austin. He is currently an Assistant Professor at The Hong Kong Polytechnic University.

\end{document}